# GATE: Graph Attention Neural Networks with Real-Time Edge Construction for Robust Indoor Localization using Mobile Embedded Devices

DANISH GUFRAN

Colorado State University, Fort Collins, CO, USA, Danish.Gufran@colostate.edu

SUDEEP PASRICHA

Colorado State University, Fort Collins, CO, USA, Sudeep@colostate.edu

Accurate indoor localization is crucial for enabling spatial context in smart environments and navigation systems. Wi-Fi Received Signal Strength (RSS) fingerprinting is a widely used indoor localization approach due to its compatibility with mobile embedded devices. Deep Learning (DL) models improve accuracy in localization tasks by learning RSS variations across locations, but they assume fingerprint vectors exist in a Euclidean space, failing to incorporate spatial relationships and the non-uniform distribution of real-world RSS noise. This results in poor generalization across heterogeneous mobile devices, where variations in hardware and signal processing distort RSS readings. Graph Neural Networks (GNNs) can improve upon conventional DL models by encoding indoor locations as nodes and modeling their spatial and signal relationships as edges. However, GNNs struggle with non-Euclidean noise distributions and suffer from the GNN blind spot problem, leading to degraded accuracy in environments with dense access points (APs). To address these challenges, we propose GATE, a novel framework that constructs an adaptive graph representation of fingerprint vectors while preserving an indoor state-space topology, modeling the non-Euclidean structure of RSS noise to mitigate environmental noise and address device heterogeneity. GATE introduces *1)* a novel Attention Hyperspace Vector (AHV) for enhanced message passing, *2)* a novel Multi-Dimensional Hyperspace Vector (MDHV) to mitigate the GNN blind spot, and *3)* an new Real-Time Edge Construction (RTEC) approach for dynamic graph adaptation. Extensive real-world evaluations across multiple indoor spaces with varying path lengths, AP densities, and heterogeneous devices demonstrate that GATE achieves 1.6× to 4.72× lower mean localization errors and 1.85× to 4.57× lower worst-case errors compared to state-of-the-art indoor localization frameworks.

## 1 INTRODUCTION

Indoor localization has emerged as a critical application in recent years, driving context-aware advancements in smart environments, including spatial robotics, location tracking in augmented/virtual reality (AR/VR), and precise emergency response systems [1]. The first documented indoor localization framework, "*The Active Badge Location System*", was proposed in 1992 [2]. This system relied on infrared (IR) signals, where stationary IR receivers detected periodic IR pulses emitted by wearable badges. However, due to the inability of IR signals to penetrate walls, the system was limited to line-of-sight localization, making it impractical for large-scale deployments. As indoor environments have grown increasingly complex, researchers have turned to artificial intelligence (AI) to improve localization accuracy [1], [3] . This shift has driven major investments from industry leaders such as Apple, Google, Microsoft, HPE, Quuppa, and Zebra, driving innovations in AI-powered localization systems [4]. As a result, the global indoor localization market was valued at $11.9 billion (USD) in 2024, with projections indicating sustained growth fueled by the increasing adoption of intelligent localization technologies [5].

Modern indoor localization primarily relies on data-driven approaches that leverage wireless radio frequency (RF) signal measurements from established protocols such as Wi-Fi, Bluetooth, Ultra-Wideband (UWB), and cellular 4G/5G networks [1], [3], [4]. Among these, Wi-Fi-based localization has gained significant traction due to its widespread availability and seamless integration with mobile devices [6]. Wi-Fi-based localization estimates a device's location by measuring Received Signal Strength (RSS) values from Wi-Fi routers or Access Points (APs) that are widely deployed in most indoor spaces today [7]. RSS values, measured in decibels relative to a milliwatt (dBm), typically range from −100 dBm (indicating a weak or undetectable signal) to 0 dBm (indicating a strong received signal). By analyzing RSS values from multiple APs, a mobile device's location can be estimated. As the device moves, RSS variations across APs enable real-time tracking, making Wi-Fi-based localization a practical solution for many applications. Despite its advantages, Wi-Fi RSS-based localization faces many challenges due to environmental interference and device heterogeneity, both of which degrade localization accuracy [7], [8]. Signal absorption, reflection, and diffraction caused by walls, furniture, and electronic devices introduce unpredictable variations in RSS values. Additionally, human movement, multipath propagation, and shadowing effects further contribute to RSS variations. Device heterogeneity exacerbates these challenges, as different mobile device manufacturers implement proprietary Wi-Fi chipsets, antenna placements, and signal processing algorithms, leading to inconsistent RSS measurements—even for devices at the same location. Prior studies have identified hardware-induced heterogeneity, caused by variations in antennas and processing units (CPU, GPU, NPU, FPGA), and software-induced heterogeneity, where vendor-specific firmware optimizations (e.g., noise filtering) alter raw RSS readings [8], [9]. Collectively, these factors significantly impact the reliability of Wi-Fi-based localization, necessitating robust techniques to mitigate environmental noise and heterogeneity-induced variations.

To address these challenges, researchers have explored various algorithms for localization with wireless signals, such as time difference of arrival (TDoA) and angle of arrival (AoA), with fingerprinting-based localization emerging as the dominant approach for indoor spaces [3], [7], [9]. The fingerprinting method constructs RSS fingerprint vectors, which are feature representations of RSS values received from multiple APs at a given location. These fingerprints are collected at multiple locations (henceforth referred to as Reference Points (RPs)) across an indoor space, creating a structured database which enables real-time tracking. In general, the fingerprinting-based localization process consists of two key phases: *1) Offline Phase*: where RSS fingerprint vectors are collected at all RPs and stored in a database, and *2) Online Phase*: where a mobile device at an unknown location captures a new fingerprint vector and compares it against the stored database to determine the closest matching RP. Despite its widespread adoption, fingerprinting-based localization remains highly sensitive to environmental noise and device heterogeneity, both of which cause substantial variations in RSS values, leading to diminished localization accuracy [10]. Simple similarity-matching approaches with database entries often yield poor results in dynamic indoor environments, which has prompted researchers to explore Machine Learning (ML) based solutions. Models such as K-Nearest Neighbors (KNN) [11], Support Vector Machines (SVM) [12], Gaussian Process Classification (GPC) [13], Deep Neural Networks (DNNs) [14], Convolutional Neural Networks (CNNs) [15], and Attention-based models [16] have shown significant improvements by learning intricate RSS patterns across RPs (as will be discussed in Section 2). However, while these models can reduce the impact of environmental noise, they fail to capture spatial relationships between RPs (the topology of the indoor path) and struggle with device heterogeneity induced errors. A fundamental limitation arises from the non-Euclidean nature of Wi-Fi fingerprinting, where noise distribution varies irregularly across the fingerprint vector due to environmental interference and device-specific variations (as will be discussed in Section 3). Yet, state-of-the-art ML models for indoor localization assume a Euclidean data structure, treating noise as uniformly distributed across the fingerprint vector [17]. This incorrect assumption leads to degraded localization performance, particularly in large and heterogeneous environments.

To overcome the limitations of traditional ML approaches, Graph Neural Networks (GNNs) have emerged as a promising solution by integrating graph theory within the ML paradigm, allowing these models to process fingerprint vectors while preserving the topology of the indoor space [18]. Unlike conventional ML models which treat fingerprint vectors as independent feature sets, GNNs explicitly model spatial relationships between RPs by representing the indoor space as a graph. In this representation, each RP serves as a node, embedded with its corresponding fingerprint vector, while edges between nodes are constructed based on spatial proximity and RSS similarity [17], [18]. This structure enables GNNs to learn spatial dependencies instead of treating RPs as isolated points, as discussed later in Section 3. As a result, GNNs significantly outperform traditional ML models, particularly in environments where device heterogeneity and environmental noise introduce RSS variations that degrade the performance of conventional approaches [18].

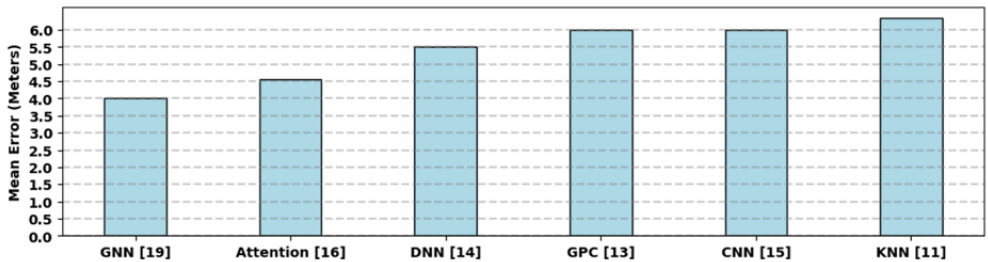

Figure 1: Localization performance comparison of GNN [19] against state-of-the-art ML methods for a 60-meter-long indoor environment. ML models evaluated include an Attention-based Neural Network [16], a fully connected Deep Neural Network (DNN) [14], Gaussian Process Classification (GPC) [13], a Convolutional Neural Network (CNN) [15], and K-Nearest Neighbor (KNN) [11].

To quantify the effectiveness of GNNs for indoor localization, we conducted real-world experiments along a 60-meter-long indoor path in a building environment, where RPs were spaced 1 meter apart, and each RP observed up to 183 APs (thus a fingerprint vector at each RP has a maximum dimension of 183). To assess device heterogeneity, RSS fingerprint data was collected using seven different mobile devices (smartphones). Further details on dataset specifications, device configurations, and building floorplans are provided in Section 5. Figure 1 presents a comparative analysis of a traditional GNN model-based solution [19] against other ML models used in state-of-the-art indoor localization solutions. Our results demonstrate that the GNN achieves the lowest localization error, outperforming KNN [11] and Attention-based models [16]— both widely used in prior works (and discussed further in Section 2). Despite their advantages, our analysis indicates that GNNs suffer from scalability issues in large indoor environments with many APs, primarily due to the "GNN blind spot problem", where conventional GNNs struggle to capture long-range dependencies in high-dimensional graphs (in the presence of high AP density per RP) and fail to model the non-Euclidean distribution of noise in RSS fingerprints [20]. Since GNNs rely on message passing to propagate node information through edges, they inherently assume a Euclidean noise structure, which unfortunately degrades performance in noisy, real-world settings. While message passing enables nodes to aggregate information from edges, it loses the relative influence of each edge on the node, making the learned representations noisy and less discriminative. This effect is further exacerbated in environments with high AP densities, where multiple APs contribute inconsistent RSS values, amplifying RSS variations and reducing localization accuracy.

To collectively address environmental noise, device heterogeneity, GNN blind spots, and non-Euclidean data modeling, in this work we introduce GATE (Graph Attention Neural Networks with Real-Time Edge Construction)—a novel GNN-based framework for robust indoor localization. Our key contributions are:

- We introduce the Attention Hyperspace Vector (AHV), which improves message passing by preserving the non-Euclidean noise structure of fingerprint vectors, ensuring more robust edge influence modeling.
- We propose the Multi-Dimensional Hyperspace Vector (MDHV), which captures an indoor topology while incorporating edge influence to mitigate the GNN blind spot problem.
- We develop a Real-Time Edge Construction (RTEC) policy, which dynamically refines graph connectivity during inference, enhancing resilience to environmental noise and device heterogeneity.
- We conduct extensive real-world experiments across diverse buildings with varying path lengths, AP densities, and multiple heterogeneous devices to evaluate the GATE framework's robustness.
- We benchmark GATE's performance against state-of-the-art GNN-based indoor localization methods, demonstrating its effectiveness in the presence of real-world RSS variations.

The remainder of this paper is structured as follows: Section 2 reviews related works in the domain of indoor localization, highlighting key advancements and approaches that have been proposed to address localization challenges. Section 3 delves into the nature of RSS fingerprint vectors, examining their inherent structure and discussing methods to represent these vectors as a graph. Section 4 introduces the GATE framework, detailing its architecture, components, and operational workflow. Section 5 presents the experimental setup and is followed by a comprehensive evaluation of GATE's performance against state-of-the-art indoor localization solutions. Lastly, Section 6 summarizes our key findings.

## 2 RELATED WORK

With the adoption of ML in indoor localization in recent years, a wide range of approaches have been proposed to make RSS fingerprinting more viable in real-world deployments. Classical ML algorithms such as K-Nearest Neighbors (KNN) [11], Support Vector Machines (SVM) [12], Gaussian Process Classification (GPC) [13], Decision Trees (DT) [21], and Non-Negative Matrix Factorization (NMF) [22] showed improvements over basic similarity-matching approaches by learning patterns in fingerprint vectors across RPs. These models served as an early validation of the feasibility of ML in fingerprinting-based indoor localization. However, they remained highly sensitive to variations in RSS due to environmental noise and device heterogeneity, as they lacked the capacity to differentiate between meaningful RSS patterns and noise artifacts. This limitation prompted a transition toward deeper ML models such as Deep Neural Networks (DNNs) [14] and Convolutional Neural Networks (CNNs) [15], which improved performance by extracting higher-level representations from fingerprint vectors. The deeper models are capable of separating RSS patterns from small-scale random noise by focusing on higher-order feature representations. Yet, in real-world environments, RSS fluctuations are not only random but also follow non-uniform noise distributions (non-Euclidean structure) that these models are unable to efficiently generalize over [17].

To better capture such variability, researchers began leveraging data augmentation techniques during training to simulate real-world noise conditions. These included synthetic noise injection strategies using architectures such as autoencoders and generative adversarial networks (GANs). Notable frameworks like SANGRIA [23], WiDeep [13], and DataLoc [24] explored several data augmentation methods to generate augmented RSS fingerprints, effectively mimicking environmental and heterogeneity-induced noise. Simultaneously, approaches like STONE [25], and STELLAR [26] incorporated contrastive learning, which relies on augmentation to learn invariant representations by contrasting noisy and clean RSS samples. These methods enriched the training distribution and offered improved resilience to moderate noise. However, despite their successes, these models continued to treat fingerprint vectors as existing in a Euclidean space, assuming noise to be uniformly distributed across all APs. None of these methods model the spatial relationships between RPs, i.e., the indoor path topology, which is critical for understanding how RSS values change with user movement and the physical layout of the indoor environment. Without incorporating this topological context, existing solutions that use deep ML models lack the spatial awareness required for high accuracy indoor localization.

To address the structural limitations of conventional deep ML models, Graph Neural Networks (GNNs) have recently emerged as a promising paradigm for fingerprinting-based indoor localization by modeling fingerprint vectors as graphs [19]. In this formulation, RPs are represented as graph nodes, and edge connections capture spatial relationships or RSS similarities between nodes. This allows GNNs to preserve the topological structure of the indoor environment, which traditional deep ML models lack. However, GNNs are still relatively new in this domain, with only a few published works. One of the earliest attempts was GNN-LOC proposed in 2021 [19]. Although this approach leveraged graph-based modeling to encode indoor topology, it lacked message passing, and therefore failed to model the influence of the edges—particularly the interference effects caused by overlapping APs in nearby RPs. This limitation led to the development of extensions like GNN-ED [27] and GCN-ED [28], which continued to use the Euclidean distance for edge construction—typically by thresholding the average pairwise Euclidean distance across fingerprint vectors—but added message passing layers to enable feature aggregation across connected nodes. GCN-ED additionally interpreted the graph as an image and applied convolutional filters to learn spatial patterns. Building on this, GNN-KNN [29] and GCN-KNN [30] used KNN for edge construction, where edges were formed based on RSS similarity. These models improved performance by incorporating message passing to learn topological dependencies and demonstrated better resilience to environmental noise. However, they still operated under the assumption of a Euclidean data structure and failed to account for non-uniform noise introduced by device heterogeneity. To further improve learning under such variability, attention mechanisms were integrated into the GNN framework. In particular, GCLoc [31] and GraphLoc [35], which are variants of Graph Attention (GAT) networks allowed each node to assign learned attention weights to its edges, effectively prioritizing informative connections and reducing the influence of unreliable or noisy nodes. However, even GCLoc and GraphLoc remains limited by its reliance on Euclidean assumptions and its vulnerability in high-density AP environments, where modeling becomes increasingly complex.

In summary, despite recent advances, GNN-based indoor localization solutions continue to underperform in environments with high AP densities (experimental evidence across varying AP densities is presented in Section 5.5). In such settings, the graph structure becomes increasingly noisy, leading to incorrect or redundant edge connections that disrupt spatial dependencies. We term this as the "GNN blind spot problem", which arises when message passing is insufficient to transmit information across nodes with high AP density, resulting in feature dilution. While the specific term "GNN blind spot" is not universally adopted, the phenomenon has been recognized in GNN literature as oversquashing [32], over-smoothing [33], and neighborhood dilution [34]. Together, these limitations highlight the need for a more advanced framework that overcomes three fundamental challenges: *1)* overcoming both environmental noise and device heterogeneity by preserving the inherently non-Euclidean structure of fingerprint vectors, *2)* maintaining indoor topology within the graph structure to ensure spatial context is retained, and *3)* addressing the GNN blind spot by enabling robust message passing even in high-density AP environments.

## 3 REPRESENTING FINGERPRINT VECTORS AS A GRAPH FOR INDOOR LOCALIZATION

In this section, we first analyze and discuss the structural properties of fingerprint vectors and how they affect the learning dynamics of conventional ML models. Next, we explore graph-based representations of fingerprint vectors, explaining how spatial and signal-based relationships can be effectively encoded to enhance localization accuracy. Lastly, we examine existing message passing algorithms in GNNs, discussing their implementations and how they contribute to the GNN blind spot challenge. The insights from this analysis inform the development of GATE, ensuring that the framework preserves the structural properties of fingerprint vectors while addressing the limitations of existing GNN based solutions.

### 3.1 Structural Analysis of Wi-Fi RSS Fingerprint Vectors: Euclidean vs. Non-Euclidean Representation

Determining whether a dataset exhibits a Euclidean or non-Euclidean structure is critical when selecting an appropriate learning algorithm. In a Euclidean paradigm, data points are assumed to lie in a space where distances can be measured using Euclidean distance (ED)—the straight-line distance between two points. This structure imposes fixed, uniform relationships between features, implying that variations in RSS values across different APs are evenly distributed, with a constant noise offset ($\delta$) applied uniformly across all dimensions of the fingerprint vector. Most ML models are inherently designed to operate within such Euclidean spaces. Each RSS value in the fingerprint vector is treated as an independent feature, and these models implicitly assume that noise—whether from environmental interference or device heterogeneity—is uniformly distributed across all APs, as shown in Figure 2 (left). However, real-world fingerprint vectors violate these assumptions. The variations introduced by environmental and device-specific factors are inherently non-uniform, meaning that some APs are affected more than others depending on physical layout, environmental noise, and device heterogeneity. This assumption degrades the generalizability of traditional ML models, limiting their effectiveness in real-world indoor localization. For example, if a fingerprint vector is denoted as $F$, as shown in Eq. (1), Euclidean models assume a constant offset $\delta$ across all APs, resulting in a transformed vector $F_{Euclidean}$, shown in Eq. (2).

$$F = [RSS_{AP:0}, RSS_{AP:1}, \ldots, RSS_{AP:N}] \quad (1)$$

$$F_{Euclidean} = [(\delta, RSS_{AP:0}), (\delta, RSS_{AP:1}), \ldots, (\delta, RSS_{AP:N})] \quad (2)$$

In reality, the $\delta$ introduced at each AP varies in a non-uniform manner. Therefore, the true $F$ follows a non-Euclidean structure, where each AP-specific offset $\delta_i$ differs, as expressed in Eq. (3):

$$F_{Non-Euclidean} = [(\delta_0, RSS_{AP:0}), (\delta_1, RSS_{AP:1}), \ldots, (\delta_N, RSS_{AP:N})], \quad (3)$$

Figure 2 (right) illustrates this effect, showing how RSS values exhibit spatially inconsistent $\delta_i$, disrupting the structured relationships assumed by Euclidean models. Hence, to accurately process $F$ a more sophisticated learning approach is required—one that can operate effectively in non-Euclidean spaces.

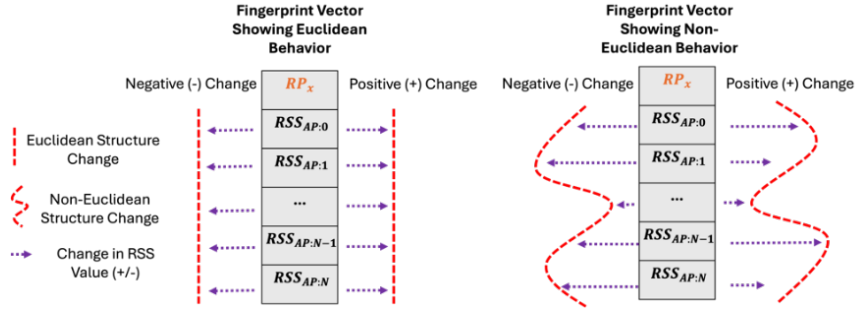

Figure 2: Fingerprint vector with Euclidean noise assumption (left) vs. real-world non-Euclidean structure (right).

### 3.2 Representing Fingerprint Vectors as a Graph

To effectively leverage GNNs for indoor localization, it is essential to construct a graph representation of fingerprint vectors $F$ that captures both the topology of the indoor environment—i.e., how RPs are spatially structured along the indoor path—and the inherent non-Euclidean relationships among them. Let $F_i$ denote $F$ at RP $i$. Unlike traditional ML models that treat $F_i$ as an independent feature vector in Euclidean space, a graph-based approach explicitly models the spatial dependencies between RPs. This enables the model to extract localization cues not just from direct fingerprint similarities, but also from the spatial context—crucial for generalization in environments affected by irregular $\delta_i$. In this representation, each RP is mapped as a node in the graph, with its associated $F$ serving as the node feature, as illustrated in Figure 3 (left). Edges between nodes encode spatial and signal-based relationships, and their structure should ideally reflect the physical layout and signal dynamics of the environment. Formally, this graph ($G$) can be defined as:

$$G = (V, E, F) \quad (4)$$

Here, $V$ denotes the set of nodes, where each node $v_i \in V$ corresponds to an RP with its associated fingerprint vector $F_i \in R^N$, where $N$ represents the number of available APs. The set of edges $E \in \{(v_i, v_j)\}$ captures relationships between RPs and can be defined based on spatial proximity or RSS similarity. The effectiveness of this graph representation hinges on the quality of the edge construction strategy. If the edges do not accurately reflect the underlying topology of the indoor environment, the graph structure fails to model meaningful spatial dependencies, ultimately degrading localization performance. Many existing GNN-based approaches employ static graph construction methods, such as KNN or ED-based RSS similarity, discussed next in Section 3.3. These methods often overlook the non-uniformity of RSS noise and treat all AP measurements as equally reliable, which can lead to incorrect node associations, linking RPs with similar RSS values but distant physical locations. Such erroneous connections distort the topology of the graph and introduce noise into the message passing process. To enable effective learning, the graph must accurately preserve local spatial relationships between nearby RPs while also maintaining enough global structure to model long-range dependencies within the indoor environment. However, achieving this balance is challenging due to the irregular nature of RSS data, and remains a key limitation in current GNN-based architectures used for indoor localization.

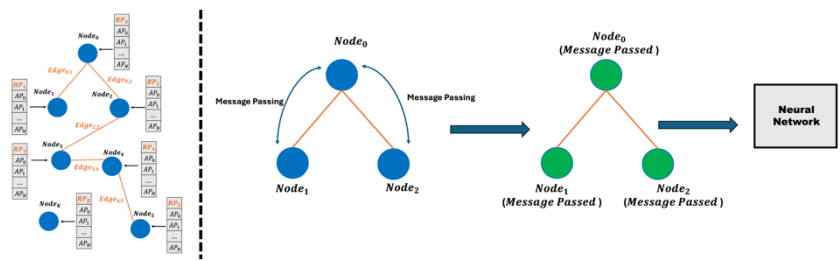

Figure 3: Graph representation of fingerprint vectors (left) and message passing between nodes (right).

Although constructing an appropriate graph structure is a critical step, it alone is insufficient. Once the graph is defined, GNNs rely on message passing to exchange information across nodes, enabling the model to refine each RP's representation based on its neighbors. However, in large-scale indoor environments with dense AP deployments, conventional message passing strategies become less effective, which we define as the GNN blind spot problem. In the next section, we analyze the working of existing message passing algorithms and motivate our design choices for GATE.

### 3.3 Edge Construction and Message Passing for Fingerprint Vectors

Once fingerprint vectors $F$ are represented as a graph, the accuracy and efficiency of indoor localization heavily depend on how edges $E$ are constructed and how information is propagated between nodes $V$ through message passing. The $E$ in a GNN define how spatial and signal-based dependencies are captured, while message passing dictates how information flows between RPs, influencing the model's ability to infer relationships between $V$. Incorrect $E$ construction can distort the non-Euclidean relationships in $F$, leading to incorrect feature propagation and degraded localization accuracy. A few $E$ construction and message passing strategies have been explored for GNN-based localization, each with distinct advantages and limitations. We analyze the three most widely used approaches next.

#### 3.3.1 Euclidean Distance-Based Message Passing

In Euclidean Distance (ED) based edge construction, nodes $V$ are connected based on the average pairwise ED between their respective $F$. The idea is that nodes with low ED values are assumed to be more similar in signal space and are therefore likely to be spatially or topologically related. A threshold $\emptyset_{ED}$ is set to filter connections, ensuring that only nodes with sufficiently similar RSS patterns are linked. The ED between two nodes $(v_i, v_j)$ is computed as:

$$ED(v_i, v_j) = \frac{1}{N} \sum_{k=1}^{N} |F_i^k - F_j^k|^2 \tag{5}$$

where $F_i^k$ denotes the RSS value from the $k^{th}$ AP in the fingerprint vector $F_i$ and $N$ is the total number of APs. Once the ED score is calculated, an edge is formed between $(v_i, v_j)$ only if their ED falls below $\emptyset_{ED}$ as shown below:

$$E = \{ (v_i, v_j) \mid ED(v_i, v_j) \leq \emptyset_{ED} \} \tag{6}$$

After edge construction, feature aggregation is performed using weighted message passing scheme. Each node updates its representation by averaging the features of its neighbors, scaled by learned weights $\alpha_{i,j}$:

$$F_i' = \sigma(W * \sum_{j \in N(V_i)} ( \alpha_{i,j} * F_j) \tag{7}$$

Here $W$ is the trainable weight matrix and $\sigma$ is a non-linear activation function such as *RELU*. Thus, ED-based edge construction connects nodes with similar RSS values, but this approach suffers from key limitations in real-world indoor localization. Due to environmental noise and device heterogeneity, two distant RPs may occasionally share similar fingerprints. As a result, this method can produce incorrect edges—either missing valid neighbors or including unrelated nodes—leading to noisy message passing and reduced localization accuracy.

### 3.3.2 K-Nearest Neighbor (KNN)-Based Message Passing

In KNN based edge construction, each node is connected to its *K* most similar neighbors, typically using ED as the similarity metric. While the same distance function is used as in threshold-based ED construction (see Eq. (5)), the key distinction lies in how edges are formed. Rather than applying a global threshold (Eq. (6)), KNN uses a local ranking strategy, forming edges to the *K* nodes with the lowest ED scores relative to each node. This ensures that each node maintains exactly *K* connections, promoting stable graph connectivity even in the presence of environmental noise or signal sparsity. Formally, an edge $(v_i, v_j) \in E$ is created only if $v_j$ lies among the top *K* nearest neighbors of $v_i$ based on the ED ranking:

$$E = \{ (v_i, v_j) \mid v_j \in k_{v_i}(ED(v_i, v_j)) \} \tag{8}$$

Once edges are formed, message passing proceeds similarly to the formulation shown in Eq. (7). KNN-based construction offers two notable advantages over global threshold-based ED: *1)* it guarantees that each node has a consistent number of neighbors, mitigating the risk of disconnected subgraphs; and *2)* it adapts locally to fingerprint distributions, making it more robust to noise variations across the environment. However, it still shares a major limitation with ED-based methods—both rely on Euclidean similarity, which does not inherently account for the non-Euclidean structure of fingerprint vectors. As a result, KNN may still form spurious edges between signal-similar but spatially distant nodes.

### 3.3.3 Attention-Based Message Passing

Unlike KNN-based edge construction, which enforces a fixed number of neighbors, or ED-based methods that rely on a global threshold, attention-based message passing learns the importance of each neighbor during training. Instead of using static edge weights, the attention mechanism assigns a learnable importance score to each edge based on the attention similarity between the node features. Given two node embeddings $F_i$ and $F_j$, the attention coefficient $\alpha_{i,j}$ is computed as:

$$\alpha_{i,j} = \frac{EXP\left(LeakyRELU\left(a^T\left[W*F_i \,\|\, W*F_j\right]\right)\right)}{\sum_{K \in N_{(V_i)}} EXP\left(LeakyRELU\left(a^T\left[W*F_i \,\|\, W*F_j\right]\right)\right)} \qquad (9)$$

where $W$ is a learnable weight matrix, $a^T$ is the learnable attention vector, and $\|$ denotes concatenation. Once the attention coefficients are computed, message passing proceeds using the same weighted aggregation described in Eq. (7), where neighbors are now weighted by $\alpha_{i,j}$. This allows the model to adaptively prioritize informative nodes and ignore noisy or less useful ones. However, despite their advantages, attention mechanisms still depend on the initial Euclidean structure, which may introduce spurious connections. In environments with dense AP distributions, they remain vulnerable to the GNN blind spot, where long-range dependencies are lost, and message passing becomes ineffective.

## 4 GATE FRAMEWORK FOR ROBUST INDOOR LOCALIZATION

### 4.1 Overview of the GATE Framework

GATE introduces a novel graph-based learning framework for Wi-Fi RSS fingerprinting that preserves both the indoor topology and the non-Euclidean structure of fingerprint vectors. Unlike traditional ML models that treat fingerprint vectors as independent features in Euclidean space, GATE first establishes edges based on the indoor topology, ensuring that spatial relationships between RPs are preserved. It then assigns attention scores to each edge, based on the similarity between node embeddings and also models the graph to preserve the non-Euclidean structure of fingerprint vectors. This combined strategy enables GATE to adaptively capture both indoor topology and signal-based dependencies, addressing key challenges such as environmental noise, device heterogeneity, and the GNN blind spot problem.

GATE operates in two key phases: an offline training phase and an online inference phase, as illustrated in Figure 4. During the offline phase, fingerprint vectors are collected at multiple RPs across the building. Each RP is treated as a node in a graph, where the total number of nodes corresponds to the number of RPs in the training set. To establish edges between nodes, GATE employs a hybrid approach that considers both indoor topology and signal-based attention scores. Edges are first formed based on indoor topology, and attention scores are then assigned to quantify the relative importance of each edge. These scores are computed using a scalar dot product between node embeddings, as detailed in Section 4.2. The number of neighbors per node is treated as a tunable hyperparameter, whose influence on localization performance is examined in Section 5.2. Nodes with higher attention scores are emphasized during training, allowing the model to strengthen meaningful spatial dependencies while suppressing weak or noisy associations that could hinder accuracy.

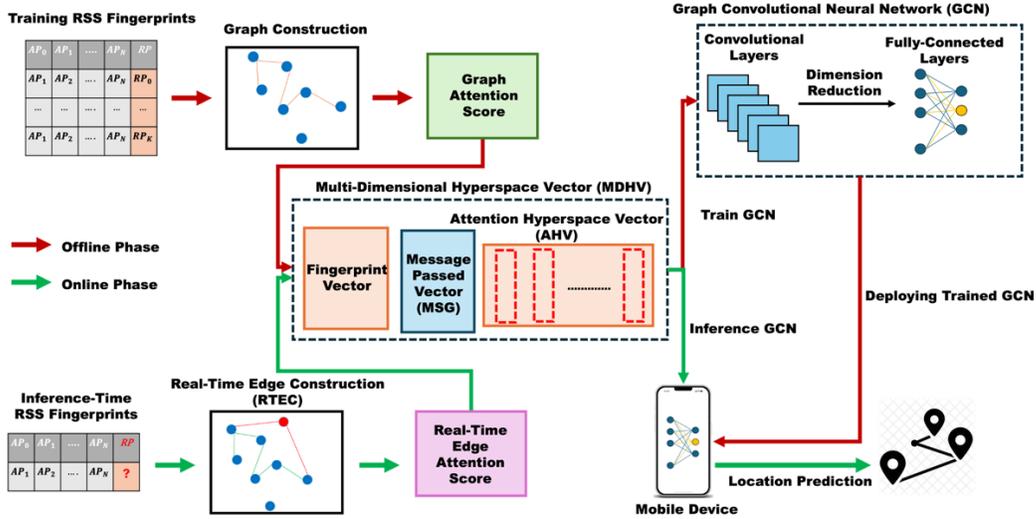

Figure 4: Overview of the GATE framework during offline training and online inference phases.

Once the graph is constructed, GATE computes a Multi-Dimensional Hyperspace Vector (MDHV) for each node, which serves as the primary feature representation for training, as shown in Figure 4. The MDHV consists of three components: *1)* the fingerprint vector of the node, representing raw RSS values, *2)* the attention-based message passing (MSG) vector, which aggregates information from connected nodes based on their edge weights, and *3)* the Attention Hyperspace Vector (AHV), a multi-dimensional tensor encoding the feature-wise influence of each neighboring node. While the MSG vector captures the overall influence of connected nodes, the AHV preserves the non-Euclidean structure by maintaining per-feature contributions, thereby accounting for the non-uniform impact of environmental noise and device heterogeneity. Together, these components enable the MDHV to encode both the influence of each edge and the fine-grained contribution of each neighbor's features, preserving spatial integrity and signal variability in a structured form. This representation enhances robustness against non-uniform distortions and directly addresses the GNN blind spot problem by ensuring that feature-level distinctions are retained even under dense AP conditions. Once the MDHVs are computed for all nodes, they are passed into a custom-designed Graph Convolutional Network (GCN) for training. We select GCNs because the MDHV structure resembles image-like spatial feature maps, making convolutional layers particularly well-suited to extract hierarchical patterns and spatial relationships. The GCN architecture, as shown in Figure 4, consists of two graph convolutional layers followed by a fully connected classification layer. The first layer captures high-order dependencies across the graph, while the second compresses the feature space to retain only the most discriminative attributes before final location prediction.

In the online phase, when a mobile device captures an RSS fingerprint at an unknown location, it is temporarily added to the graph as a new node without altering the trained GCN. GATE applies Real-Time Edge Construction (RTEC), described in Section 4.5, to compute attention scores between the new node and existing nodes, assigning edges to the most relevant ones. The number of edges is kept fixed to match the training configuration and maintain dimensional consistency. The MDHV for the new node is then computed using the same approach as in training, ensuring compatibility with the learned graph structure and allowing the GCN to accurately predict the most likely RP.

### 4.2 GATE Attention Score Mechanism

The attention score mechanism in GATE is designed to quantify the similarity between fingerprint vectors and determine the strength of connections between nodes in the graph. Given two RSS fingerprint vectors, $F_i$ and $F_j$, corresponding to two RPs, the attention score is computed using the scalar dot product attention. This is defined as:

$$a_{i,j} = \frac{F_i \cdot F_j}{||F_i|| \cdot ||F_j||} \tag{10}$$

where $a_{i,j}$, represents the attention score between node $i$ and node $j$, and the dot product captures the alignment between the two fingerprint vectors. The denominator normalizes the score, ensuring that the attention values remain bounded and comparable across different RPs. This formulation ensures that fingerprint vectors with similar RSS distributions receive higher attention scores, effectively modeling the spatial and signal-based relationships between locations. Once attention scores are computed for all connected nodes, they are used as weights in a weighted averaging scheme to perform message passing. The message passing process aggregates feature information from neighboring nodes while ensuring that more relevant nodes contribute more significantly to the target node's representation. The aggregated message $MSG_i$ for node $i$ is computed as:

$$MSG_i = \sum_{j \in N(i)} (A_{i,j} \cdot F_j) \tag{11}$$

where $N(i)$ represents the set of neighboring nodes connected to node $i$. This message passing step ensures that the spatial dependencies between fingerprint vectors are encoded into the node representation, improving the model's ability to generalize across different environments. The computed message vector $MSG_i$, along with the original fingerprint vector $F_i$, is then integrated into the Multi-Dimensional Hyperspace Vector (MDHV) for further computations, as discussed next.

### 4.3 Multi-Dimensional Hyperspace Vector (MDHV) Computation

The Multi-Dimensional Hyperspace Vector (MDHV) serves as a representation for each node in the GATE framework, ensuring that the spatial and signal-based dependencies within the fingerprint vectors are effectively captured. The MDHV consists of three components: *1)* the fingerprint vector $F_i$, which contains the raw RSS values from multiple APs, *2)* the message-passed $MSG_i$ vector, which aggregates information from connected nodes based on attention-weighted averaging, and *3)* the Attention Hyperspace Vector (AHV), a multi-dimensional tensor encoding the feature-wise influence of neighboring nodes. The AHV plays a crucial role in capturing the non-uniform impact of each neighboring node's features on the target node. While traditional attention mechanisms assign a single weight per node, the AHV expands this concept by computing feature-wise attention, ensuring that the contribution of each RSS feature is dynamically adjusted. The AHV is computed as:

$$AHV_{i,j} = \frac{F_i \odot F_j}{||F_i|| \cdot ||F_j||} \tag{12}$$

where $\odot$ denotes element-wise multiplication, ensuring that each feature within $F_i$ is compared individually against $F_j$, a neighboring node. This formulation ensures that RSS features with higher correlation across nodes receive greater weightage during message passing, while irrelevant or noisy features contribute minimally. The AHV is structured as a multi-dimensional tensor, where each column corresponds to a connected node and each row represents an individual attention score for each RSS feature, with the total dimensionality dictated by the number of connected edges per node. By integrating the fingerprint vector, message-passed vector, and the AHV, the MDHV preserves both the local and global dependencies within the graph while maintaining the non-Euclidean structure of the fingerprint vectors. This structured representation is then supplied as input to the Graph Convolutional Network (GCN), allowing the model to effectively learn spatial and signal-based patterns crucial for accurate indoor localization.

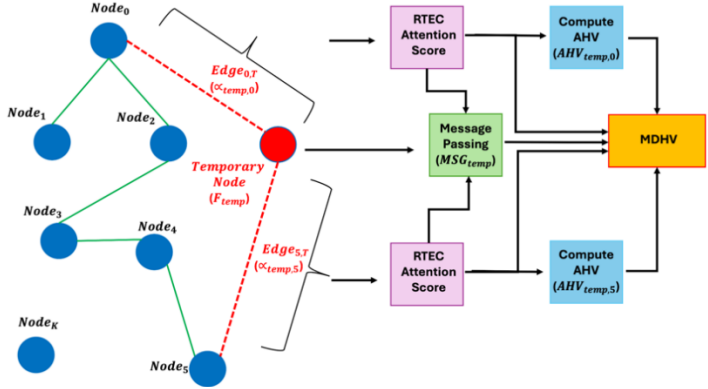

Figure 5: Overview of the RTEC algorithm during the online phase.

### 4.4 Real-Time Edge Construction (RTEC) Algorithm

The Real-Time Edge Construction (RTEC) algorithm plays a crucial role in integrating new fingerprint vectors dynamically into the GATE framework during the online phase. Unlike static graph structures used in conventional GNN-based models, RTEC dynamically constructs edges for a temporary node (e.g., red node in Figure 5). When a new RSS fingerprint vector $F_{temp}$ is captured from an unknown location, RTEC treats it as a temporary node in the graph. To establish meaningful connections with existing nodes, RTEC computes attention scores between $F_{temp}$ and the fingerprint vectors of pre-existing RPs using the same scalar dot-product attention mechanism applied in the offline phase. The attention score between the temporary node and an existing node $i$ is given by:

$$\propto_{temp,i} = \frac{F_{temp} \cdot F_i}{||F_{temp}|| \cdot ||F_i||} \tag{13}$$

where $\propto_{temp,i}$ measures the similarity between the newly observed fingerprint and the stored fingerprint of RP $i$. The number of edges connected to the temporary node is kept fixed, as determined during the offline phase, to maintain dimensional consistency in the graph structure. Once the edges are determined, message passing is performed using attention-weighted averaging, aggregating information from the selected neighbors, as shown in Figure 5. The message-passing output $MSG_{temp}$ is computed as:

$$MSG_{temp} = \sum_{i \in N_{temp}} (\propto_{temp,i} \cdot F_i) \qquad (14)$$

where $N_{temp}$ represents the set of selected neighbors for the temporary node. The MDHV is then constructed using the same methodology as in the offline phase, integrating the fingerprint vector, the message-passed vector, and the AHV, which captures the feature-wise impact of each connected node. Finally, the constructed MDHV is passed into the pre-trained GCN, which predicts the most probable RP associated with the fingerprint vector. RTEC allows GATE to perform real-time localization by dynamically selecting neighbors based on signal similarity, computing attention-aware feature aggregations, and maintaining structural consistency with the trained graph.

## 5 EXPERIMENTAL RESULTS

### 5.1 Experimental Setup

We evaluated the performance of the GATE framework using real-time data collection to closely emulate practical deployment scenarios. Experiments were conducted across five distinct buildings, with floorplans in each differing in the number of RPs, AP densities, and spatial topologies—factors that introduce varying and challenging environmental noise characteristics. To additionally consider the impact of device heterogeneity, we conducted experiments with seven commercially available smartphones with diverse hardware and software configurations. For each RP, up to five fingerprint vectors were collected for training, while one fingerprint vector was reserved for testing. The spatial layouts of the buildings and the specifications of the devices used are illustrated in Figure 6.

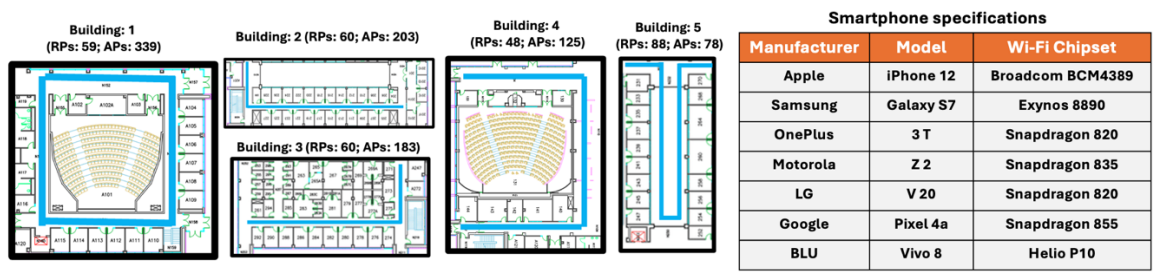

Figure 6: Spatial layouts of the building floorplans (left) and mobile device specifications (right).

We selected five buildings (Building 1–5), each chosen for its unique structural and environmental characteristics. The test paths span various shapes and lengths, ranging from 48 to 88 meters. Building 1 exhibits the highest AP density, with up to 339 visible APs per RP and 59 RPs, resulting in a fingerprint vector of shape (59, 339). It features a mix of wood and concrete infrastructure and is surrounded by computers in large open areas. In contrast, Building 5 has the lowest AP density, with only 78 APs per RP but the highest RP count of 88, forming a fingerprint vector of shape (88, 78). This building contains laboratory equipment with several different electronic and mechanical devices. Buildings 2 and 3 each have 60 RPs, but differ in AP densities: Building 2 has 203 APs per RP (60, 203), while Building 3 has 183 APs per RP (60, 183). Both buildings are newer constructions with a blend of metallic and wooden structures, containing open spaces, bookshelves, and heavy metallic objects. Building 4 represents the shortest path, with 48 RPs and the second-lowest AP density of 125 (48, 125). It includes multiple small office rooms with a combination of metal, wood, and glass partitions. To ensure consistent data collection across all environments, a fixed granularity of 1 meter was maintained between adjacent RPs. For training, we used the Motorola Z2 device, while all seven smartphones were employed for testing to comprehensively evaluate generalization under device heterogeneity.

The training process for GATE is designed to optimize model efficiency while ensuring high accuracy indoor localization. We configure the GCN with a learning rate of 0.001 and utilize the sparse categorical cross-entropy loss function. The first CONV layer in the GCN is configured with filters equal to the size of the MDHV, applying a kernel size of 1. The second CONV layer reduces the feature dimension to *H* (GCN compression value), where we explore the impact of varying *H* values in Section 5.2. Both CONV layers employ the *RELU* activation function, followed by a fully connected layer with neurons set to the number of RP classes for the corresponding building floorplan. The final classification output uses the *Softmax* activation function. The model was trained for 1,000 epochs, with a total of 51,542 trainable parameters, and a compact model size of 604 KB, making it feasible for efficient deployment on resource-limited devices, such as wearables and smartphones. For performance evaluation, we measure localization error using the absolute distance metric. The error for each prediction is computed as the absolute difference between the predicted RP ($RP_{Pred}^i$) and the ground truth RP ($RP_{True}^i$) in meters, as shown below:

$$Error_{Device} = \frac{1}{N}\sum_{i=1}^{N}|RP_{Pred}^i - RP_{True}^i| \qquad (15)$$

where $N$ is the total number of RPs, $RP_{Pred}^i$ is the predicted RP class for RP $i$, and $RP_{True}^i$ is the ground truth RP for RP $i$. The mean localization error for a given test device ($Error_{Device}$) is calculated as the average of these absolute differences across all RPs within a building floorplan.

### 5.2 Sensitivity Analysis: GCN Compression Size and Number of Edges per Node

In this section, we show results of explorations on the sensitivity of GATE to two core hyperparameters: the GCN compression size (*H*) and the number of edges per node (*NB*), as shown in Figure 7. The compression size *H* represents the dimensionality reduction between the first and second convolutional (CONV) layers in the GCN and is varied from 0% (no compression) to 90% (retaining only 10% of the original feature dimensions), plotted along the Y-axis. *NB* defines the percentage of nodes connected to each node based on attention scores, and is varied from 1% (sparse connections) to 100% (fully connected graph), plotted along the X-axis. The Z-axis in each plot shows the mean localization error (in meters) averaged over seven testing devices. This setup allows us to systematically study how GATE responds to changes in feature compression and edge density, and more importantly, where GNN blind spots begin to emerge.

Starting with compression size, we observe that Buildings 1, 2, and 3—each characterized by high AP densities and hence high-dimensional fingerprint vectors—exhibit a relatively flat localization error surface across *H* values. This indicates that GATE is robust to feature compression in such environments, with only minor improvements around the 40% to 60% compression range. The model can afford to drop some feature dimensions without significant degradation because the original vectors already capture rich spatial and signal characteristics. In contrast, Buildings 4 and 5, which have lower AP densities and therefore smaller fingerprint vectors, show noticeable increases in error beyond 60% compression. In these environments, overly aggressive compression leads to the loss of key discriminative features within the MDHV, resulting in reduced localization performance. Overall, the aggregated performance across all buildings confirms that GATE remains largely stable with respect to *H*, with 50% compression yielding the lowest mean errors.

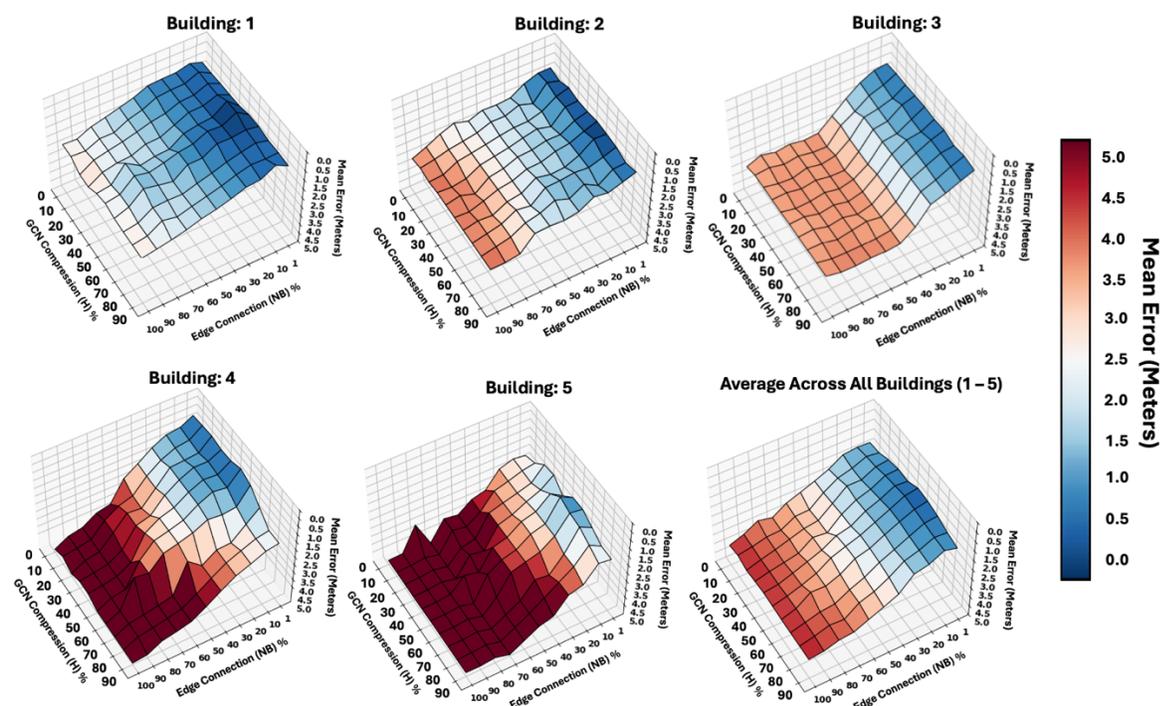

Figure 7: Sensitivity of GATE to GCN compression size (*H*%) and edge connectivity (*NB*%) for each building floorplan.

The number of edges per node (*NB*) has a more pronounced impact on the GATE framework's behavior. Across all buildings, we observe a clear trend: increasing *NB* leads to progressively higher localization errors. This degradation is due to the GNN blind spot problem, where excessive connectivity introduces representational noise, especially in dense graphs. While AHV provides per-feature attention weighting to mitigate this effect, it cannot entirely resolve the dilution of spatial specificity when irrelevant nodes are aggregated. Interestingly, GATE shows strong resilience to this blind spot problem in Buildings 1 and 2, where despite very high AP densities, errors remain below 2 meters up to *NB* = 60%. This is a direct result of the MDHV structure, which leverages both AHV and MSG to model inter-node relevance and feature-wise contrasts effectively. Even in Building 3, GATE maintains stable performance up to *NB* = 40%, which is notably higher than what traditional GNNs can tolerate in similar settings (see experimental analysis in Section 5.6). In contrast, the blind spot problem becomes more apparent in Buildings 4 and 5, with performance deterioration observed at much lower *NB* values—around 40% and 20%, respectively. These environments lack the high-dimensional input space needed for AHV and MSG to construct more representative data embeddings, making them more sensitive to edge noise. Despite this, GATE still outperforms conventional baselines by a significant margin, as will be discussed in Section 5.6. The sixth, aggregated plot provides a generalized view: maintaining *NB* at or below 10% and setting *H* to approximately 50% yields consistently low localization error across all environments. This configuration is the one we consider for subsequent analysis as it allows GATE to significantly mitigate the GNN blind spot problem in indoor localization.

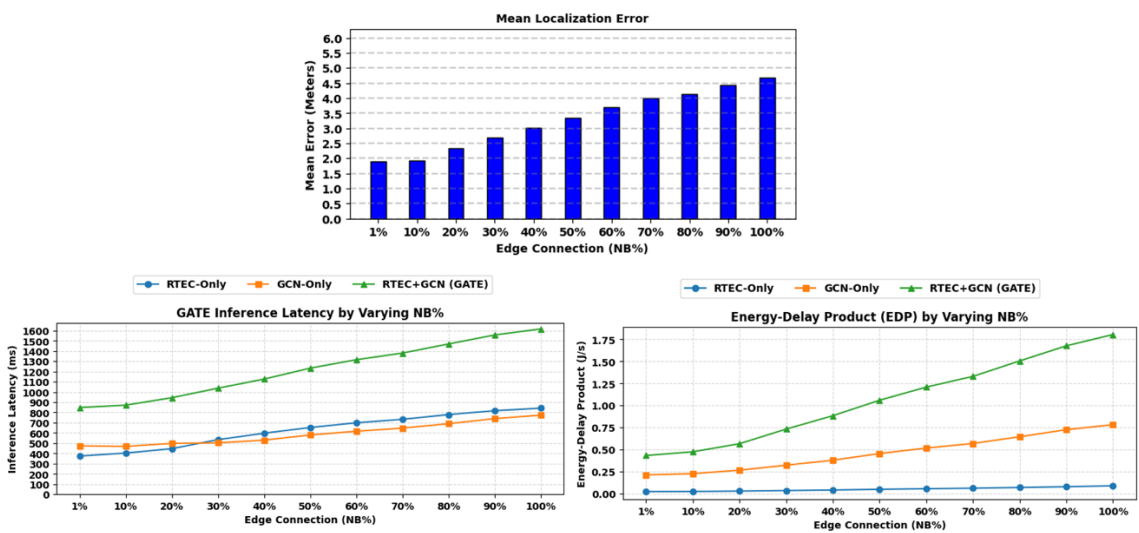

Figure 8: Impact of edge connectivity (NB%) on GATE's localization accuracy, inference latency, and energy-delay product (EDP) across all mobile devices and buildings.

To evaluate the real-world deployability of GATE, we analyze its performance across all mobile devices and buildings presented in Figure 6, focusing on how NB% impacts both localization accuracy and on-device efficiency. We consider two critical deployment metrics: inference latency—the time required to generate location predictions—and energy-delay product (EDP)—a measure of the energy consumed to perform inference. These metrics are evaluated under three configurations: *1) RTEC-Only*, where we capture the time and energy required to compute attention-based edge connections and construct the MDHV; *2) GCN-Only*, which measures the cost of performing inference on the MDHV; and *3) RTEC+GCN*, which reflects the total end-to-end latency and energy consumption of the full GATE pipeline. Figure 8 presents the results: the top bar plot shows the mean localization error (in meters) for each NB%, averaged over all mobile devices and buildings. The bottom-left and bottom-right line plots show the corresponding average inference latency (in milliseconds) and EDP (in joules per second), respectively. In all plots, the X-axis represents NB%, while the Y-axis denotes the respective performance metrics.

In the *RTEC-Only* configuration, inference latency remains low and stable for NB values between 1% and 20%, but increases progressively beyond this point, reaching a maximum of 842 ms at NB = 100%. This trend is expected, as higher NB values lead to an increased number of attention score calculations and message-passing operations required to construct the MDHV. Despite this rise in latency, the EDP remains relatively flat across all NB settings, ranging from 0.06 J/s at NB = 1% to 0.11 J/s at NB = 100%. This suggests that although execution time grows with higher edge connectivity, energy consumption does not rise proportionally. The efficiency of RTEC stems from its use of lightweight scalar dot-product attention operations, which are less compute-intensive than the dense matrix operations typical of convolutional layers. As a result, RTEC remains highly energy-efficient and well-suited for execution on mobile devices.

In the *GCN-Only* configuration—which is executed after the MDHV are constructed—we observe that inference latency remains nearly constant for NB values between 1% and 40%, with only a slight increase beyond that. GCN inference is consistently faster than RTEC, as the GCN does not need to compute attention scores individually for every node during inference; instead, it operates on the pre-computed MDHV (from RTEC), which is highly optimized on modern mobile device hardware. However, the EDP increases noticeably at higher NB values, especially beyond 30%. This increase is driven by the growing size of the MDHV in denser graphs. Larger input dimensions result in more intensive feature-wise computations within each convolutional layer, leading to higher energy consumption despite relatively stable runtime. When evaluating the full *RTEC+GCN* pipeline, we find that total inference latency increases from 848 ms at NB = 1% to 1616 ms at NB = 100%, while EDP rises from 0.5 J/s to 1.11 J/s over the same range. Despite this growth, both metrics remain within acceptable thresholds for real-time inference on modern mobile devices, demonstrating GATE's practicality for real-time deployments. In summary, our analysis identifies NB values between 1% and 10% as the optimal operating range for real-world use. Within this range, GATE consistently delivers localization errors below 2 meters, end-to-end inference latency under 1 second, and EDP below 0.6 J/s across all devices and buildings. These findings affirm that GATE is not only accurate and robust but also energy-efficient for deployment on resource-constrained mobile devices.

### 5.3 Sensitivity Analysis: Effects of Fingerprint Samples per RP on Heterogeneity Resilience

In this section, we evaluate the impact of varying the number of fingerprint samples collected per RP, and how this influences the GATE framework's ability to generalize across heterogeneous devices. Specifically, we vary the number of training samples per RP from 1 to 5, in increments of 1, while keeping the test configuration fixed at one fingerprint sample per RP. Figure 9 presents the results across all seven smartphones, with separate plots for each building and an aggregated plot that captures the overall trend. The X-axis denotes the seven devices, and the Y-axis denotes the mean localization error in meters. As illustrated in Figure 9, increasing the number of training samples per RP consistently reduces the mean localization error and substantially lowers device variance—quantified as the difference between the highest and lowest localization error among devices. Our analysis reveals those buildings with higher AP densities—specifically Buildings 1, 2, and 3—exhibit lower worst-case localization errors. For instance, with five samples per RP, these buildings achieve mean localization errors of 1.36, 1.43, and 1.84 meters, respectively across all devices. In contrast, Buildings 4 and 5, which have lower AP densities, observe higher errors of 2.39 and 2.41 meters, respectively across all devices. When examining the aggregated results across all buildings, training with only one sample per RP yields the highest device variance: the mean localization error ranges from 3.84 meters (worst-performing device) to 2.35 meters (best-performing device), reflecting a variance of 1.49 meters. In comparison, training with 2, 3, 4, and 5 samples per RP results in reduced variances of 1.15, 1.07, 0.5, and 0.25 meters, respectively. Notably, training with five samples per RP achieves the lowest device variance, representing a 1.83× improvement in heterogeneity resilience compared to the single-sample case.

This enhanced resilience is attributed to the higher RSS fingerprint variability obtained from multiple samples per RP. When more samples are used during training, the resulting RSS values introduce subtle but meaningful variability for each node, which subtly improves the variability of the MSG vector. This broader RSS distribution allows the AHV to capture fine-grained feature-wise variability across node edges, leading to a more diverse and robust MDHV representation. Consequently, the GATE framework becomes better equipped to generalize to devices with differing signal characteristics. Empirically, this results in a mean localization error of 1.98 meters across all test devices and buildings when using five samples per RP. Based on this analysis, we fix the number of training samples per RP to five for all subsequent experiments.

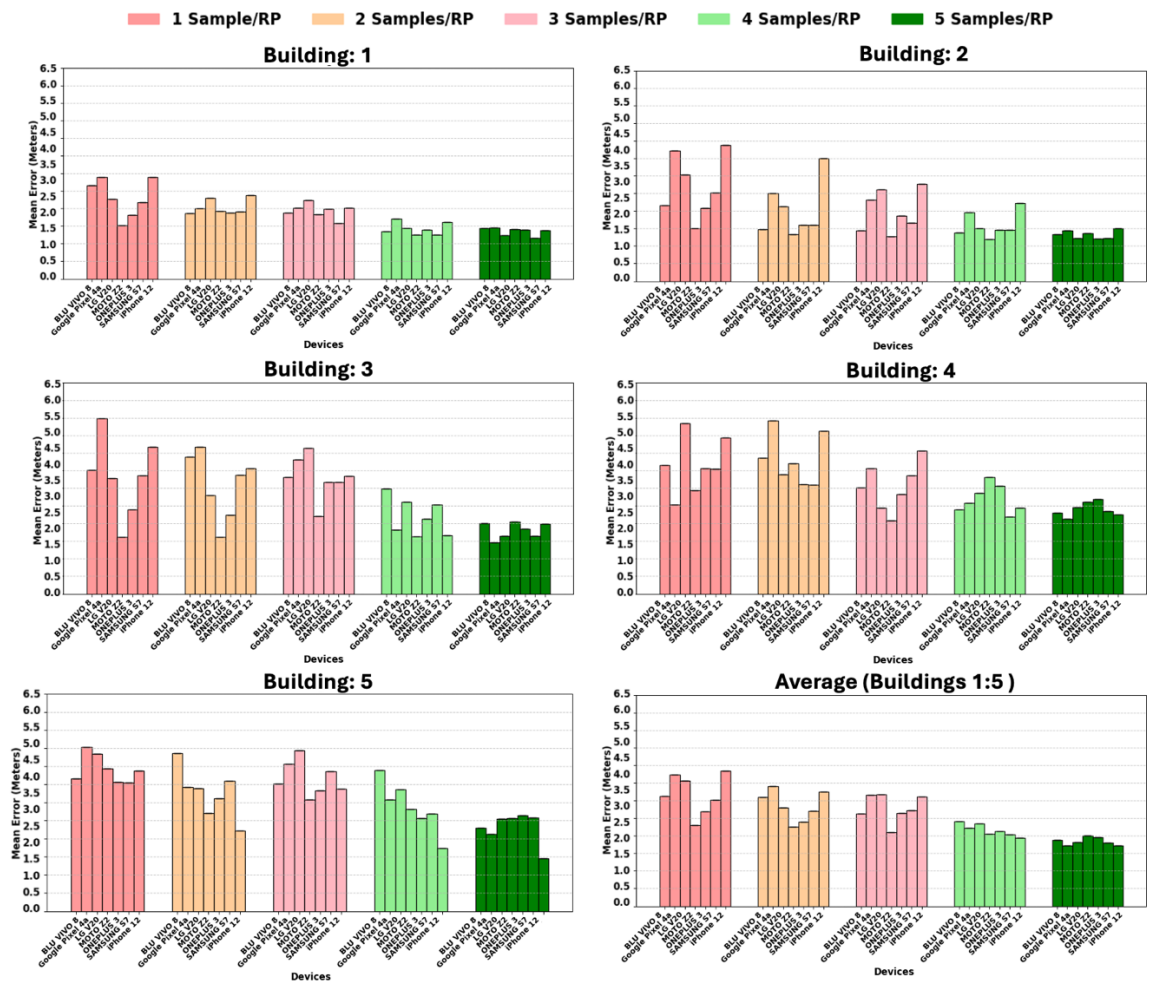

Figure 9: Sensitivity of GATE to the number of fingerprint samples per RP during training.

## 5.4 Component-Wise Analysis of GATE: Impact on Worst-Case Errors and Heterogeneity Resilience

Next, we assessed the contributions of each major component in the GATE framework by comparing four architectural variants: *1) GATE-No-MSG* which removes the MSG vector and constructs the MDHV using only the fingerprint vector and AHV during both training and RTEC-based inference, *2) GATE-No-AHV* which excludes the AHV while retaining the fingerprint vector and MSG vector during both training and RTEC-based inference, *3) GATE-No-MDHV* which removes both MSG and AHV, relying solely on the fingerprint vector as input and disabling RTEC during inference, and *4) GATE-Full* which uses all three components—fingerprint vector, MSG, and AHV—within the MDHV during both training and RTEC-based inference. Figure 10 presents the results across all seven devices, averaged over the five building floorplans. The X-axis denotes the testing devices, and the Y-axis denotes the mean localization error in meters. The result for each device is shown as a box plot, where the red line represents the mean error across all RPs and building floorplans, the upper whisker indicates the worst-case error across all RPs for that device, and the lower whisker shows the best-case error across all RPs for that device.

From Figure 10, we observe that *GATE-No-MDHV* performs the worst among all variants, exhibiting the highest worst-case error and the highest susceptibility towards heterogeneity. The Samsung S7 records the largest outlier at 8.3 meters, while the device variance—defined as the difference in mean localization error between the best and worst performing devices—peaks at 2.25 meters, ranging from 4.75 meters on the BLU Vivo 8 to 2.5 meters on the Moto Z2. This highlights the importance of the MDHV in managing environmental noise and device-induced signal variability. *GATE-No-AHV* results in the second-highest worst-case error of 7.15 meters (Google Pixel 4a) but shows improved resilience with a lower device variance of 1.1 meters, suggesting that the presence of the MSG vector helps capture inter-node context and stabilizes performance across devices. However, the absence of AHV limits the model's ability to account for feature-wise irregularities, leaving it vulnerable to localized noise. In contrast, *GATE-No-MSG* reduces the worst-case error slightly to 5.8 meters but suffers from high device variance (2.17 meters), ranging from 3.27 meters (BLU Vivo 8) to 1.1 meters (Moto Z2). This indicates that while AHV helps suppress extreme errors, omitting MSG weakens the model's generalization across devices. *GATE-Full* achieves the best results overall, with the lowest worst-case error of 3.2 meters (OnePlus 3T) and minimal device variance of 0.25 meters. These findings confirm that the combination of MSG and AHV within MDHV enables our framework to learn both edge-level influence and feature-level contributions, preserving the non-Euclidean structure and achieving robustness towards device heterogeneity and environmental noise.

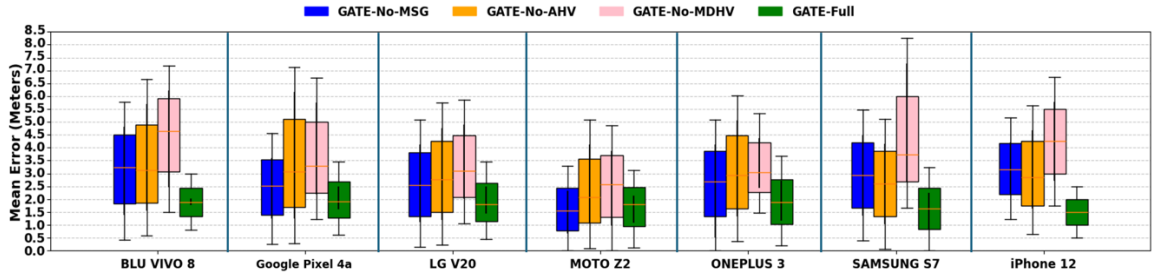

Figure 10: Component-wise analysis of different GATE variants.

## 5.5 Impact of Varying AP Densities on GATE Variants: Performance Across Truncated Fingerprint Vector Size

We also investigated how the performance of the four GATE variants—*GATE-No-MDHV*, *GATE-No-AHV*, *GATE-No-MSG*, and *GATE-Full*—varies with the dimension of the fingerprint vectors per RP. We simulated environments with limited AP visibility by progressively truncating each fingerprint vector, randomly dropping a percentage of RSS values per RP while keeping the graph connectivity intact. This experiment assesses the robustness of each GATE variant in situations where fewer APs are visible. In Figure 11, the X-axis denotes fingerprint vector truncation percentage varied and the Y-axis denotes the mean localization error (in meters), averaged across all RPs, buildings, and devices.

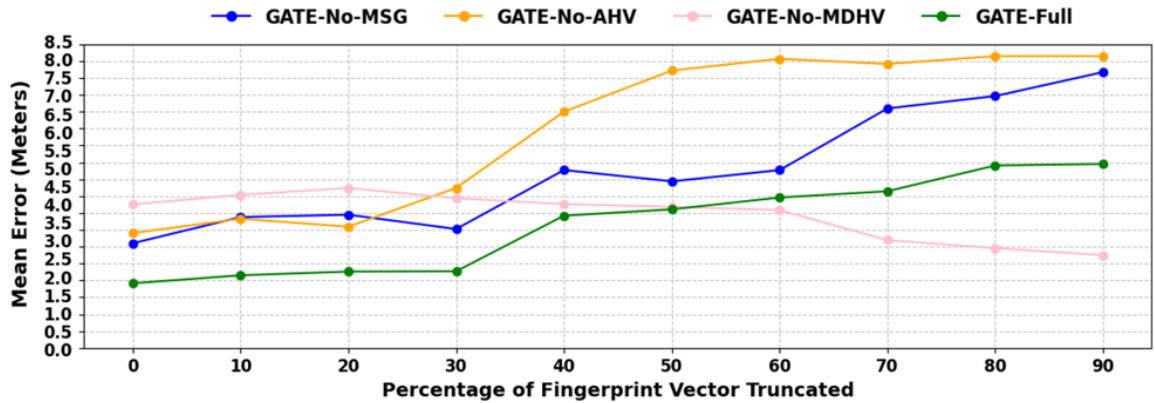

Figure 11: Impact of varying AP densities (fingerprint vector truncation) on localization accuracy across GATE variants.

The *GATE-No-AHV* variant shows the sharpest increase in mean localization error as the fingerprint vector is truncated. Performance remains stable up to 20% truncation, but errors rise significantly between 30% and 60%, plateauing at higher truncation levels. This behavior highlights the importance of AHV in capturing per-feature contributions during message passing. Without AHV, the MSG vector aggregates larger noise from neighboring nodes, especially when fewer RSS features are available, reducing discriminative capacity. The *GATE-No-MSG* variant performs slightly better but still shows marked error increases beyond 30% and again around 60%. While AHV offers some local feature sensitivity, its effectiveness declines with reduced input dimensionality as the attention scores rely on feature variation to assign meaningful weights, when dimensions are heavily truncated, this variation diminishes, making the scores less discriminative. The *GATE-Full* variant yields lower errors across all truncation levels compared to *GATE-No-AHV* and *GATE-No-MSG*, with only mild increases around the 40% and 80% marks, and maintains a bounded error of 5.5 meters at 90% truncation. Interestingly, the *GATE-No-MDHV* variant shows a counterintuitive trend: localization error decreases as truncation increases beyond 50%. This behavior likely arises because lower-dimensional inputs simplify the learning task for models without contextual aggregation, especially in simpler layouts or low-variability settings. However, such configurations lack robustness and fail under heterogeneous or complex environments with higher AP densities. These findings suggest that in scenarios with small fingerprint vectors (i.e., low AP density), simpler GNNs/GCNs may suffice, as message passing introduces noise that outweighs its benefits. However, as AP density increases and the environment becomes more complex, message passing alone becomes unreliable due to the GNN blind spot problem. In these cases, the MDHV becomes critical, preserving both contextual and per-feature signal variation, thereby effectively mitigating the GNN blind spot problem in high-dimensional settings.

### 5.6 Comparison of GATE Variants with State-of-the-Art Indoor Localization Frameworks

Lastly, we compare the performance of GATE and its variants against state-of-the-art GNN-based localization frameworks from prior work. Figure 12 summarizes the results, where the X-axis denotes the evaluated frameworks and the Y-axis indicates mean localization error in meters, averaged across five building floorplans and seven mobile devices. In each box plot, the upper whisker denotes the worst-case error, the lower whisker denotes the best-case error, and the orange line denotes the mean error. As can be observed in Figure 12, *GATE-Full* achieves the lowest mean localization error of 1.98 meters and the lowest worst-case error of 3.5 meters across all building floorplans and devices, outperforming all current state-of-the-art frameworks. The second-best performance comes from *GATE-No-MSG*, which lacks the message passing (MSG) vector. In comparison, *GATE-Full* yields 1.38× lower mean error and 1.61× lower worst-case error, demonstrating the importance of MSG for device heterogeneity resilience. The GCLoc [31] and GraphLoc [35] frameworks from prior work rank third and fourth due to their use of Graph Attention (GAT) based architectures. However, their lack of AHV makes them unable to capture the non-Euclidean feature structure of fingerprint vectors, especially in high AP density environments. *GATE-Full* surpasses GCLoc and GraphLoc with 1.6× to 1.8× lower mean error and 1.8× to 1.9× lower worst-case error respectively. *GATE-No-AHV* performs similarly to GCLoc in terms of worst-case error but has a slightly higher mean error, with *GATE-Full* achieving 1.73× lower mean error in comparison. Traditional models such as GNN-KNN [29], GCN-KNN [30], GNN-ED [27], GCN-ED [28], STONE [25], STELLAR [26], SANGRIA [23], DNN [14], CNN [15], and KNN [11] show significantly higher mean and worst-case errors. These frameworks are limited by fixed graph structures and Euclidean assumptions, making them vulnerable to signal variability and the GNN blind spot effect. *GATE-Full* achieves 2.26×, 2.24×, 2.48×, 3.31×, 3.03×, 3.4×, 3.9×, 4.12×, 4.42×, and 4.72× lower mean errors and 2.62×, 4.00×, 3.22×, 4.05×, 2.56×, 2.61×, 3.22×, 4.28×, 4.28×, and 4.57× lower worst-case errors respectively. These results indicate that integrating both the MSG vector and AHV within the MDHV not only preserves the non-Euclidean structure of RSS fingerprints but also mitigates noise propagation during message passing—particularly in high-dimensional fingerprint environments—effectively mitigating the GNN blind spot effect and enhancing overall localization performance.

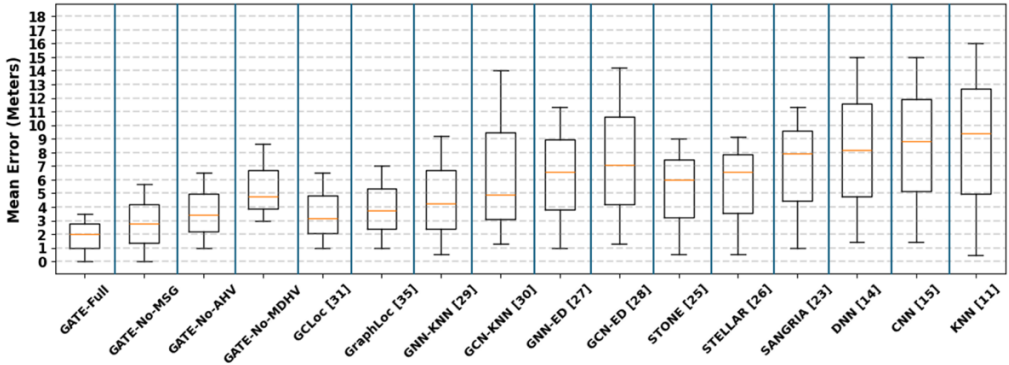

Figure 12: Evaluation of the GATE variants compared to state-of-the-art indoor localization frameworks.

Next, we evaluate the real-world deployability of GATE and the baseline frameworks on the mobile devices detailed in Figure 6. We analyze several key hardware metrics presented in Table 1—averaged across all seven devices and five buildings—to understand each model's suitability for mobile deployment. These metrics include: 1) Inference Latency: The time required for a single location prediction, measured in milliseconds (ms). Lower latency is essential for real-time applications such as indoor localization. 2) Model Size: This is the total memory footprint of the trained model, reported in kilobytes (KB). Models with smaller memory footprints are easier to deploy on mobile devices with limited storage and memory constraints. 3) Energy-Delay Product (EDP): This metric captures energy efficiency by measuring the energy consumed per prediction, measured in Joules per second (J/s). Lower EDP indicates the model is energy-efficient on battery-powered devices. 4) FLOP (×10³): This measures the total number of floating-point operations required for a single inference, reflecting the model's computational complexity. A higher FLOP value typically corresponds to models being more computationally demanding.

Table 1: Comparison of hardware performance metrics (averaged across all seven devices and five buildings) for GATE and baselines.

| Model | Mean Error (Meters) | Inference Latency (ms) | Model Size (KB) | EDP (J/s) | FLOP (×$10^3$) |
|---|---|---|---|---|---|
| GATE-No-MSG | 2.75 | 836 | 599.3 | 0.521 | 576.8 |
| GATE-No-AHV | 3.45 | 799 | 590.4 | 0.493 | 319.6 |
| GATE-No-MDHV | 4.75 | 734 | 509.3 | 0.444 | 271.5 |
| **GATE-Full** | **1.98** | **871** | **604.0** | **0.543** | **714.2** |
| GCLoc [31] | 3.19 | 904 | 1404.8 | 0.53 | 786.4 |
| GRAPHLOC [35] | 3.75 | 1034 | 2234.1 | 0.75 | 982.3 |
| GNN-KNN [29] | 4.25 | 934 | 934.8 | 0.61 | 644.4 |
| GCN-KNN [30] | 4.9 | 1077 | 1034.0 | 0.65 | 796.9 |
| GNN-ED [27] | 6.6 | 698 | 498.7 | 0.50 | 349.0 |
| GCN-ED [28] | 7.1 | 754 | 511.3 | 0.52 | 414.7 |
| SANGRIA [23] | 7.89 | 437 | 1774.3 | 0.47 | 262.2 |
| STONE [25] | 6.0 | 1301 | 5667.1 | 1.01 | 1301.0 |

| STELLAR [26] | 6.6 | 1554 | 8024.0 | 1.20 | 1864.8 |
| DNN [14] | 8.2 | 855 | 204.2 | 0.40 | 282.1 |
| CNN [15] | 8.8 | 1231 | 636.3 | 0.73 | 726.2 |
| KNN [11] | 9.4 | 1531 | 1224.6 | 0.93 | 688.9 |

Starting with *GATE-Full*, this architecture achieves a balance between accuracy and hardware efficiency. As shown in Table 1, it achieves the lowest mean localization error of 1.98 meters—substantially outperforming all baselines—while maintaining a moderate latency of 871 ms and an EDP of 0.543 J/s. Its compact model size of 604.0 KB ensures easy deployment on mobile devices with limited storage. The relatively higher FLOP value ($714.2 \times 10^3$) reflects the added computational steps required by the integrated MSG vector and AHV modules, which captures the influence of edges and maintain the non-Euclidean RSS structures. This moderate increase in computational workload is a deliberate trade-off that enhances robustness against device heterogeneity and environmental noise. Comparatively, *GATE-No-MSG*, *GATE-No-AHV*, and *GATE-No-MDHV* achieve slightly lower latencies (836 ms, 799 ms, and 734 ms) and EDPs (0.521 J/s, 0.493 J/s, and 0.44 J/s) due to the absence of the respective modules; however, they suffer from increased mean errors (2.75, 3.45, and 4.75 meters), highlighting the importance of retaining both components for consistent performance.

Examining the baselines reveals key insights: graph attention-based (GAT) frameworks like GCLoc [31] and GraphLoc [35] exhibit significantly larger model sizes (1404.8 KB and 2234.1 KB, respectively) because they lack the RTEC module for selective node filtering, forcing them to store the entire graph during inference. This results in *GATE-Full* being 2.32× to 3.69× more compact while also achieving lower latency and EDP, as shown in Table 1. GraphLoc's multi-headed attention further inflates its FLOP value ($982.3 \times 10^3$), EDP (0.75 J/s), and latency (1034 ms), while GCLoc, despite a slightly lower EDP (0.53 J/s), suffers from higher mean errors due to its inability to efficiently curate message-passing paths without RTEC. Other graph-based baselines like GNN-KNN [29], GCN-KNN [30], GNN-ED [27], and GCN-ED [28] require lower FLOP ($349.0–796.9 \times 10^3$) and achieve lower EDP (0.50–0.65 J/s) but consistently higher errors (4.25–7.1 meters) due to the lack of the AHV module, which is essential for modeling feature-wise node influence for device heterogeneity. Traditional baselines like DNN [14], CNN [15], and KNN [11] show lower model sizes (204.2–636.3 KB) and FLOP values ($282.1 \times 10^3–726.2 \times 10^3$) but still suffer from higher errors (8.2–9.4 meters), as they cannot capture the non-Euclidean structure of RSS fingerprints effectively. STONE [25], STELLAR [26], and SANGRIA [23] stand out with extremely large model sizes (1774.3–8024 KB), very high latencies (437–1554 ms), and EDPs exceeding 1.0 J/s, rendering them impractical for mobile deployments. In summary, *GATE-Full* consistently achieves sub-2-meter mean errors, inference latency of less than 1 second, and efficient EDP of less than 0.6 J/s, across all buildings and mobile devices, demonstrating robust and practical hardware deployment on mobile devices.

## 6 CONCLUSIONS

In this paper, we introduced GATE, a novel graph-based indoor localization framework designed to: *1)* mitigate both environmental noise and device heterogeneity by preserving the inherently non-Euclidean structure of fingerprint vectors, *2)* maintain indoor topology within the graph structure to ensure spatial context is retained, and *3)* overcome the GNN blind spot problem by enabling robust message passing even in high-density Wi-Fi AP environments. GATE integrates attention-based message passing and the Attention Hyperspace Vector (AHV) into its Multi-Dimensional Hyperspace Vector (MDHV) representation, while employing Real-Time Edge Construction (RTEC) during inference to dynamically form meaningful graph structures that closely represent the characteristics of the indoor space. This enables GATE to maintain spatial relationships between indoor locations, encode feature-level signal variation, and preserve the non-Euclidean structure of fingerprint vectors. Extensive experiments across diverse indoor environments and heterogeneous devices demonstrate that GATE outperforms state-of-the-art indoor localization frameworks, achieving 1.6× to 4.72× lower mean localization errors and 1.85× to 4.57× lower worst-case errors.